\documentclass[conference]{IEEEtran}
\pdfoutput=1
\usepackage[hyphens]{url}
\usepackage{hyperref}
\hypersetup{breaklinks=true, colorlinks,allcolors=blue}
\usepackage[backend=biber, style=ieee]{biblatex}

\IEEEoverridecommandlockouts
\usepackage{amsmath,amssymb,amsfonts}
\usepackage{algorithmic}
\usepackage{graphicx}
\usepackage{textcomp}
\usepackage[table,xcdraw]{xcolor}
\usepackage{balance}
\usepackage{enumitem}
\usepackage{mathtools}  % for cartesian product symbolb bigtimes
\usepackage{orcidlink}  % for orcid in author list if wanted

\def\BibTeX{{\rm B\kern-.05em{\sc i\kern-.025em b}\kern-.08em
    T\kern-.1667em\lower.7ex\hbox{E}\kern-.125emX}}

\begin{document}

\newcommand{\comments}[1]{}

\title{Utilizing Large Language Models to Synthesize Product Desirability Datasets }

\author{John D. Hastings\,\textsuperscript{†,*}, Sherri Weitl-Harms\,\textsuperscript{‡}, Joseph Doty\,\textsuperscript{§,*}, Zachary J. Myers\,\textsuperscript{††,*}, Warren Thompson\,\textsuperscript{‡‡,*}

\thanks{\textsuperscript{†}Email: john.hastings@dsu.edu}
\thanks{\textsuperscript{‡}Department of Computer Science, Design \& Journalism (CSDJ), Creighton University, Omaha, NE, USA. Email: sherriweitlharms@creighton.edu}
\thanks{\textsuperscript{§}Email: joseph.doty@trojans.dsu.edu}
\thanks{\textsuperscript{††}Email: zachary.myers@trojans.dsu.edu}
\thanks{\textsuperscript{‡‡}Email: warren.thompson@trojans.dsu.edu}
\thanks{\textsuperscript{*}The Beacom College of Computer \& Cyber Sciences, Dakota State University, Madison, SD, USA.}

}

\maketitle

\begin{abstract}
This research explores the application of large language models (LLMs) to generate synthetic datasets for Product Desirability Toolkit (PDT) testing, a key component in evaluating user sentiment and product experience. Utilizing gpt-4o-mini, a cost-effective alternative to larger commercial LLMs, three methods, Word+Review, Review+Word, and Supply-Word, were each used to synthesize 1000 product reviews. The generated datasets were assessed for sentiment alignment, textual diversity, and data generation cost. Results demonstrated high sentiment alignment across all methods, with Pearson correlations ranging from 0.93 to 0.97. Supply-Word exhibited the highest diversity and coverage of PDT terms, although with increased generation costs. Despite minor biases toward positive sentiments, in situations with limited test data, LLM-generated synthetic data offers significant advantages, including scalability, cost savings, and flexibility in dataset production.  
\end{abstract}


\section{Introduction}

Market demand is high for valuable and efficient systems and algorithms that can give concise advice or conclusions from big datasets~\cite{Liu}. A major challenge of information extraction is a frequent lack of sufficiently labeled data for training machine learning algorithms due to labor intensiveness and cost of acquiring labels~\cite{Bhavya}. Generating synthetic datasets is often quite useful, particularly for testing purposes in numerous areas of computing, including artificial intelligence, data mining, data visualization and software engineering~\cite{Mendonça}. The sizes of samples, rules, and attributes of the synthetic datasets can be readily adjusted to meet the needs of evaluating various learning algorithms \cite{Liu}. With synthetic data, attributes such as missing values, dimensions, outliers, data format and type, trends, and patterns can be controlled \cite{redpath}. Synthetic data is controlled by the researcher, whereas real data is messy and depend on the data collection approaches, such as survey data. Obtaining data can be a relevant problem because it may have privacy concerns, have associated costs, and require a prolonged time to acquire~\cite{manzano,Goncalves}. \citeauthor*{bolon}~\cite{bolon} found that synthetic datasets were useful because they offered a controlled environment for learning algorithms to evaluate classifiers. 

Synthetic data applications, or data generators, work by manipulating data's descriptive information through category sets, probability distribution functions, mathematical formulas, and other generators \cite{manzano}. \citeauthor*{popic}~\cite{popic} used synthetic data generation in application testing to show the pros and cons of the surveyed techniques by highlighting the intended usage of the applications and the system architectures. Liu \cite{Liu} created a rule-driven framework for generating synthetic datasets with customizable attributes and rules to evaluate decision tree algorithms. Related works propose data synthesizers for testing machine learning tools \cite{lin,Mendonça} and social graph generation \cite{Humski}. 

There has been rapid growth in the application of LLMs to solving numerous problems, including generating test and training data in research. \citeauthor*{schick2021generating}~\cite{schick2021generating} used GPT2-XL to generate synthetic datasets from scratch which, without the cost of manually labeling or fine-tuning a large model, were then used to train smaller more efficient models which exhibited strong performance in standard benchmarks compared to expensive models trained on real data. Bhavya \cite{Bharadwaj} leveraged LLMs to address the challenge of low resources for information extraction by using LLMs to generate pseudo labels and generate synthetic data. \citeauthor*{anaby2020not}~\cite{anaby2020not} describes a GPT2 fine-tuned on a small existing dataset which can then be used to synthesize new labeled sentences with the augmented datasets producing improved classifier performance. In \citeauthor*{Zhuowan}~\cite{Zhuowan}, LLMs demonstrated strong reasoning abilities as automatic data annotators, generating question-answer annotations for chart images.

This paper introduces novel research which investigates the synthesis of datasets for analyzing product desirability, specifically within the framework of the Product Desirability Toolkit (PDT), an area currently lacking in available datasets. PDT is detailed in the following section. The primary goal is to determine if these synthetic datasets can provide a scalable, cost-effective option in scenarios where real data is limited. The following questions guide the study:
\begin{enumerate}[label=\textbf{RQ\arabic*}:,left=1.0em]%, labelsep=0.5em]
\item What methods might be used to to produce synthetic PDT test data using an LLM? 

\item For the methods, how well does the synthetic data align with a target sentiment score? 

\item How diverse is the text in the synthetic data?

\item What are the costs to produce such data?

\end{enumerate}
The sections that follow detail the background, methodology, results, discussion, and future work.

\section{Background}
According to ISO 9241-11 \cite{iso924111}, usability is  ``the extent to which a product can be used by a user to achieve a goal with effectiveness, efficiency, and satisfaction.''  
Measuring usability is complex because it is intrinsic to the system or object under evaluation \cite{Carrabina}. In product development, understanding implicit user sentiment is crucial for creating products that truly appeal to their intended audience, especially in user-centered design (UCD) \cite{Marina}. 

Mapping user needs onto design specifications through sentiment analysis with deep learning tools \cite{WANG2018145} has promise. However, current methods are limited to document-level sentiment classification which is unable to capture attribute-level information \cite{HAN2021114604}. Additionally, these methods require manually labeled data for training, and often classify sentiment into predefined categories which have limited implications for UCD designers \cite{HAN2021114604}. In general, the supervised nature of most sentiment classification approaches limits their practical as they require extensive manual data labeling and annotation for training \cite{HAN2021114604}.

Understanding desirability is important for improving and marketing products, but is difficult to measure \cite{Zhang2018}. Concepts like ``enjoyment", ``fun" or a product's desirability for purchase or use are not effectively captured by traditional usability studies~\cite{Benedek}. To address situations in which sentiment data or user review data is lacking, tools such as surveys and the Microsoft Product Desirability Toolkit (PDT)~\cite{Benedek,benedek2002a} can be used to evaluate user experiences. Measuring desirability along with usability offers a layer of qualitative impact of the system, hence providing a much better picture of the user experience \cite{Carrabina}.

The PDT is recognized as a valuable qualitative tool for evaluating user experience, and satisfaction for products \cite{Barnum2010, Barnum, Booth2013EndUserEO, Hastings, Li2014, Tullis, Veral, Weitl}. It aims to ``understand the illusive, intangible aspect of desirability resulting from a user’s experience with a product'' \cite{Barnum}. However, while the PDT excels at gathering rich qualitative data, it lacks inherent quantitative abilities \cite{Sauro, Veral}. 

The PDT asks users to select five adjectives from a given set that best describe  their feelings about the experience along with providing an optional explanation of their word choices. By gathering this group of word/explanation pairs, the approach is designed to capture rich, qualitative data about user experiences and perceptions. 

The advantages of using the PDT include ``1) it aims to avoid a bias toward the positive found in typical questionnaires (e.g., it has been found that if a respondent thinks that a survey intends to assess the quality of a product, they are likely to provide more positive answers about quality) and 2) it is able to more effectively uncover constructive negative criticisms in the guided interview'' \cite{Hastings}. The PDT is described as the closest tool that uses ``psychometric theory to create a user experience (UX)-relevant measure of product or service desirability'' \cite{Sauro}. The design of the PDT prompts users to tell insightful stories of their experience as they explain their word choices~\cite{Barnum2010} and provides a rich set of qualitative data related to the user's implicit desirability of the product in question \cite{Tanja1}, and only takes about 5 minutes to administer \cite{Barnum}. 

\citeauthor*{GretzelFesenmaier}~\cite{GretzelFesenmaier} suggests that a UCD methodology is needed to effectively elicit embedded and implicit knowledge while remaining applicable to a larger sample of respondents. Similarly, Volo \cite{Volo} highlighted the need for a UCD method capable of unobtrusively accessing individuals’ experience while minimizing investigator and selection biases in its translations, measurements and analyses. These requirements demand an approach that is structured and controlled yet also flexible and open-ended~\cite{packer}. A major barrier to UCD is the relative lack of formal mechanisms to translate individual user ``voices'' into the design of distinct product attributes that reflect divergent preferences~\cite{Salvador}. Devising intelligent systems that can identify users' unique needs at scale and translate them into attribute-level design feedback and recommendations is essential for effective UCD processes \cite{HAN2021114604}. 

\citeauthor*{weitl2024analyzing}~\cite{weitl2024analyzing} addressed the gap between qualitative sentiment data and quantitative analysis of PDT data for UCD by applying recent LLMs  to PDT data. The findings showed that the LLMs were able to quantitatively measure product desirability at the same level that matched the expert labeled and annotated PDT qualitative user responses, providing promise for a new method for understanding desirability.

The results from \cite{weitl2024analyzing} show a promising novel method to quantitatively measuring implicit user desirability. The combination of using the PDT and LLM sentiment analysis limits investigator and selection biases as suggested by \cite{Volo}, while the method is controlled and structured but also open-ended and flexible as suggested by \cite{packer}. The question of whether or not their method of using LLMs for quantifying user desirability from PDT data addresses the scaling up issues noted by \cite{GretzelFesenmaier, HAN2021114604} remains unanswered. Unfortunately, available PDT datasets are often small, such as n=29 in \cite{Lim}, n=10 in \cite{Imler}, n=50 in \cite{Hastings2022}, and n=56 in \cite{Hastings}. This is likely due to challenges inherit in data collection from users as discussed above. 

Before the method from \cite{weitl2024analyzing} can be efficiently analyzed in its scaling to a larger sample of respondents, large PDT datasets are needed. Generating synthetic PDT data will be useful in answering that question. 

\comments{
This research investigates the use of three distinct LLM-based methods to generate synthetic PDT software product datasets that align with target sentiments. Key assessment areas include sentiment alignment, textual diversity, and cost efficiency. %Findings indicate that
The primary goal is to determine if these synthetic datasets can provide a scalable, cost-effective option in scenarios where real data is limited. The following questions guide the study:
\begin{enumerate}[label=\textbf{RQ\arabic*}:,left=1.0em]%, labelsep=0.5em]
\item What methods might be used to to produce synthetic PDT test data using an LLM? 

\item For the methods, how well does the synthetic data align with the target sentiment score? 

\item How diverse is the text in the synthetic data?

\item What are the costs to produce such data?

\end{enumerate}
The sections that follow detail the methodology, present results, discuss their implications, and outline directions for future work.
} % comments

\section{Methodology}\label{methodology}

\subsection{Synthesizing Data (RQ1)}
In this research, experiments utilized the gpt-4o-mini model~\cite{gpt4omini}, accessed via the OpenAI API~\cite{openAIapi}, to generate synthetic PDT datasets. Although \cite{weitl2024analyzing} effectively employed gpt-4o for sentiment tasks, this research utilized gpt-4o-mini as a significantly more cost-effective alternative for generating large datasets. Specifically, gpt-4o-mini's API cost is 6\% of gpt-4o's, which makes it a better fit for large-scale data generation, assuming reasonable performance. A dataset size of 1000 hypothetical software product reviews was chosen to ensure adequate coverage of the original PDT word set~\cite{benedek2002a}, which contains 118 words in total.

Three methods were tested to generate these hypothetical PDT datasets:

\begin{enumerate}
\item \textbf{Word+Review}: A random target sentiment score between 0.0 and 1.0 (where 0 is the most negative sentiment and 1 is the most positive sentiment) was provided to gpt-4o-mini, along with a list of 10 randomly selected PDT words. The gpt-4o-mini LLM was tasked with selecting an appropriate word based on the target sentiment and generating a corresponding product review. This process was repeated 1000 times to achieve a uniform coverage of sentiments. The method could be tailored in future applications if specific sentiments (e.g., more negatively or positively skewed) are required.

\item \textbf{Review+Word}: Because Word+Review tended to produce product reviews that reused the selected word, this approach aimed to reduce that bias by first generating the hypothetical product review in line with the random target sentiment, then selecting a word that matches the meaning of the review. This process was repeated 1000 times.

\item \textbf{Supply-Word}: Because the first two methods sometimes picked words not included in the list of valid words, and did not use all 118 PDT words, this method randomly selected one of the 118 words. The word was supplied to gpt-4o-mini, which was asked to score the sentiment expressed by the word, and produce a hypothetical software product review in line with that target score. This process was repeated 1000 times. Note: although this method didn't supply a target score, by randomly supplying words across the PDT word set, it was expected to get a reasonable coverage of all possible sentiments.
\end{enumerate}

\begin{table}[!htbp]
\caption{PDT Synthesis Prompts by Method}
\label{tab:llm_prompts}
  \centering
%\resizebox{\linewidth}{!}{%
%\begin{tabular}{|p{0.09\textwidth}|p{0.85\textwidth}|}
\begin{tabular}{|p{0.22\linewidth}|p{0.68\linewidth}|}
\hline
\textbf{Method} & \textbf{Prompt} \\ \hline

Word+Review& ``For a hypothetical ''+ product + ``, you will produce hypothetical survey data that will be comprised of the respondent picking a word from the following comma separated list that describes their experience with the product, and then providing an explanation for the word choice. I will give you a sentiment score between 0.0-1.0 and for that sentiment number, you will select a word and produce a human like comment. Words: '' + word-list  \\ \hline

Review+Word& ``I will give you a sentiment score between 0.0-1.0. For that sentiment number, you will produce a hypothetical product review for a hypothetical ''+ product + `` that matches the sentiment, and then pick a word from the following comma separated list that best matches the meaning of the review. Words: '' + word-list \\ \hline

Supply-Word& ``A hypothetical ''+ product + `` has been described as: '' + word + ``. Provide a sentiment score for the word between 0.00-1.00 (two decimal places) based on your implicit understanding of sentiment. For the sentiment score, produce a hypothetical product review that captures that sentiment and is appropriate for the chosen word. The review doesn't necessarily need to include the chosen word and should capture the complex nuance of sentiment produced by a human.'' \\ \hline

\end{tabular}
%}
\end{table}

Related to \emph{RQ4}, the time and tokens consumed while generating the PDT were captured.

\subsection{Assessing the Data}
In order to measure the quality of the datasets, they were assessed according to how well the sentiment of the generated text aligned with the target scores, the diversity of text found within the sets, and the coverage of the PDT word list.

\subsubsection{Assessing Alignment (RQ2)}
%For each of these approaches, a
After generating the datasets, the word/review pairs were scored using approaches described in \cite{weitl2024analyzing}, which had determined that LLMs more accurately score implicit sentiment of PDT data versus traditional sentiment analysis approaches. For Word + Review, initially gpt-4o-mini was instructed to score the word, and then adjust that base score according to the sentiment found in the text of the review. The prompt for this approach is shown as `Base+Adjust' in Table \ref{tab:scoring_prompts}.

The gpt-4o-mini LLM sometimes struggled with the `Base+Adjust' approach and would greatly adjust negative reviews upward when it didn't make sense to do so. For example, for the target score of 0.10, the word ``Undesirable'', and the review ``I found my experience with the product to be quite disappointing. It didn't meet my expectations, and many features seemed lacking or ineffective. Overall, it left me feeling frustrated rather than satisfied.'' The gpt-4o-mini LLM scored the word as 0.10, which seems appropriate. However, based on the review. it then incorrectly adjusted the score to 0.80 with the explanation ``The word `undesirable' has a low original sentiment score, as it typically conveys a negative connotation. However, given the context of a disappointing experience with a product, its adjusted score reflects a stronger negative sentiment. The explanation reinforces feelings of frustration and dissatisfaction, aligning with the adjustment.'' The gpt-4o-mini LLM produced 57 instances with such adjustments greater than or equal to 0.50.

\begin{table}[!thbp]
\caption{PDT Scoring Prompts}
\label{tab:scoring_prompts}
  \centering
%\resizebox{\linewidth}{!}{%
%\begin{tabular}{|p{0.09\textwidth}|p{0.85\textwidth}|}
\begin{tabular}{|p{0.20\linewidth}|p{0.68\linewidth}|}
\hline
\textbf{Approach} & \textbf{Prompt} \\ \hline
Base+Adjust & ``I will give a line containing a word choice followed by an explanation for the choice. For this line, provide a sentiment analysis score for each word between 0.00-1.00 (to two decimal places) where is 0.00 is a completely negative sentiment and 1.00 is a completely positive sentiment, and then an adjusted score for the word based on the explanation. Include your confidence in the accuracy of that score (low, medium, high). Additionally, provide a carefully crafted contextual explanation for the sentiment score that is related to the meaning of the text. Please provide your response in a text-based csv format on one line, with columns for the word, original score, adjusted score, confidence, and explanation. Please do not provide any other response aside from the csv formatted data.'' \\ \hline

Complete & ``I will give a word describing a user's experience with a product followed by a full review. Based on your implicit understanding of sentiment, provide a sentiment score for the word and review between 0.00-1.00 inclusive (to two decimal places) where is 0.00 is a completely negative sentiment and 1.00 is a completely positive sentiment. Include your confidence in the accuracy of that score (low, medium, high). Additionally, provide a carefully crafted contextual explanation for the sentiment score that is related to the meaning of the text. Please provide your response in a text-based csv format on one line, with columns for the word, sentiment score, confidence, and explanation. Please do not provide any other response aside from the csv formatted data.'' \\ \hline
\end{tabular}
%}
\end{table}

Some effort was made to help gpt-4o-mini better understand the `Base+Adjust' approach (which it handled well in most cases), but in an effort to keep prompt sizes small for cost reasons -- and given that gpt-4o and Claude were known to handle this approach well \cite{weitl2024analyzing}, and that this research does not aim to establish that gpt-4o-mini can properly score text using this specific approach -- ultimately, all three methods were scored by asking gpt-4o-mini to analyze the review collectively based on the word and review. The prompt for this approach is shown as `Complete' in Table \ref{tab:scoring_prompts}. The gpt-4o-mini LLM exhibited better behavior with this approach, and it was then used across all three PDT datasets.

Because differences between the target and evaluated scores could be influenced either at the PDT generation phases (with larger differences being caused by text that is misaligned with the target score) or in the scoring phase (with sentiment score being misaligned with the actual sentiment of the text), Supply-Word was scored with both gpt-4o-mini and gpt-4o in order to examine the differences in the scoring phase of the alignment calculations -- these methods of scoring are referred to as \textit{Supply-Word-Mini} and \textit{Supply-Word-4o}, respectively. 

After scoring each PDT word/review pair with the `Complete' prompt using gpt-4o-mini, an absolute difference between the target and evaluated scores was calculated. 

\subsubsection{Scoring for Text Diversity (RQ3)}
Truly human representative text should be relatively diverse, so the datasets were scored for text diversity to see if any synthesis method might be better than another in terms of diversity. Each of the three datasets were scored based on the techniques described in \cite{shaib2024standardizing} using their python diversity scoring package which includes:
\begin{enumerate}
\item \textit{Compression ratio (CR)}. A simple measure calculated by dividing the size of the original text by the size of the compressed text. A higher score indicates greater redundancy, and thus less diversity, of the text.
\item \textit{Part-of-speech compression ratio (CR-POS)}. Measures the redundancy of part-of-speech tag sequences by compressing them. A higher score indicates more repeated syntactic patterns, and thus less diversity in the text's structure.
\item \textit{Homogenization Score: ROUGE-L (HS)}. Measures the similarity between pairs of texts based on the longest common subsequences. A higher score indicates more overlap and thus less diversity between the texts.
\item \textit{N-gram diversity score (NDS)}. Calculates the ratio of unique n-grams to total n-grams in the text. A higher score indicates greater diversity, with fewer repeated sequences of words.
\end{enumerate}

\citeauthor*{shaib2024standardizing}~\cite{shaib2024standardizing} reported that compression ratio is highly correlated with other standard text diversity metrics, and in particular found CR-POS to be quite informative. Based on their recommendations informed by results and the available scoring options in their scoring package, each PDT dataset was scored using: CR, CR-POS, HS, and NDS with 4 as the maximum n-gram length.

\section{Results \& Discussion}

\subsection{Synthesized Data (RQ1)}
The three methods described in Section \ref{methodology} were used to produce the corresponding synthetic PDT datasets, each with 1000 hypothetical software reviews. These datasets are publicly available on Zenodo~\cite{hastings2024synthetic}. Table \ref{tab:dataset_samples} shows sample data produced by each method. Each of the three methods were fully capable of generating synthetic data with some slight differences in terms of alignment, cost, and diversity which are discussed in the following subsections.

\begin{table*}[!htbp]
\caption{Sample of PDT Synthetic Data from Each Method}
\label{tab:dataset_samples}
  \centering
%\resizebox{\linewidth}{!}{%
%\begin{tabular}{|p{0.09\textwidth}|p{0.85\textwidth}|}
\begin{tabular}{|p{0.11\linewidth}|p{0.03\linewidth}|p{0.10\linewidth}|p{0.67\linewidth}|}
\hline
\textbf{Method} &\textbf{Score} & \textbf{Word} & \textbf{Review} \\ \hline
Word + Review & 0.6 & High quality & I chose 'high quality' because the product consistently delivers good results and meets my expectations. The features are well-designed, and the user experience feels polished and professional. Overall, it has a strong value that I appreciate.'' \\ \hline

Word+Review& 0.3 & Overwhelming & I found the interface to be quite overwhelming. There are too many features packed into one place, which makes it hard to navigate and figure out what I need to do. It feels like the design could benefit from a simpler, more streamlined approach that balances functionality with ease of use. \\ \hline

Review+Word& 0.8 & Essential & Game-Changer for My Workflow! I've been using this software for a few months now, and I can confidently say it has transformed the way I manage my tasks. The interface is user-friendly, making it easy to navigate and find features quickly. The reporting capabilities are robust, allowing me to generate insightful data without a headache. Customer support has been remarkable too—every time I needed help, they responded promptly and effectively. It's definitely an essential tool for anyone looking to boost productivity. Highly recommended! \\ \hline

Review+Word& 0.1 & Unrefined & I had high hopes for this software, but it has been a major disappointment. The user interface is clunky and confusing, making it difficult to navigate. Not only that, but it frequently crashes, losing my progress and causing unnecessary frustration. I expected a lot more from a product that promised efficiency. Overall, it feels like a rough draft rather than a finished tool, and I can't recommend it to anyone looking for reliability. \\ \hline

Supply-Word& 0.75 & Familiar & I recently started using this software, and I have to admit, it's like a breath of fresh air. From the moment I launched it, I felt at ease, as if I had used it for years even though it was my first time. The user interface is smooth and intuitive, making navigation a breeze. I was able to dive right into my work without wasting time learning complicated functions or button placements. However, while the familiarity is comforting, I do wonder if it could benefit from some innovative features that set it apart from other products I’ve used in the past. Overall, the charm of this software lies in its simplicity, and I appreciate that it allows me to focus on my tasks without distraction. Definitely a solid choice for anyone who values usability. \\ \hline

Supply-Word& 0.45 & Inconsistent & I recently started using this software, and I have to say, my experience has been a bit of a rollercoaster. There are features I absolutely love – the interface is user-friendly, and some tools are quite powerful. However, I've encountered several bugs that have made me question its reliability. For example, just yesterday, I noticed that a function I used daily stopped working altogether without any updates or warning. It’s frustrating because, one moment, the software feels polished and effective, and the next, it falls short. I genuinely hope the developers can iron out these issues because I see a lot of potential here, but right now, it leaves me feeling uncertain about whether to fully commit or explore other options. \\ \hline

\end{tabular}
%}
\end{table*}

Although this research prompted gpt-4o-mini only for software product reviews, the approach could be easily used for PDT dataset production for any product with minor adjustments to the prompt by simply telling it to give reviews for a different product. The prompt could be further adjusted to have the reviews look at particular aspects of the product.

\subsubsection{Word Choices}
To be truly representative, each PDT dataset should include a reasonable coverage of each word in the PDT word list. Supply-Word used all 118 PDT as they were selected programmatically at random. However, neither Word+Review nor Review+Word included all words. Given that LLMs are trained on human text, it's not surprising that they could be biased toward selecting certain words. Word+Review used 111 of the 118 words.  The top five word choices were: Confusing: 53, Frustrating: 35, Empowering: 33, Valuable: 25, and Effective: 24. The approach also incorrectly generated two words (Unhelpful, Inspirational) not found in the supplied word list. In the first instance, gpt-4o-mini selected `Unhelpful' in response to a target sentiment of 0.1 and with no negative words in the list of 10 PDT words. In the second case, `Inspirational' was selected when `Inspiring' was in the PDT list.

Review+Word used 108 of the 118 PDT words. The top five word choices were: Frustrating: 43, Dated: 34, Confusing: 31, Valuable: 29, and Poor quality: 26.

In terms of generating valid PDT data, if utilizing Word+Review and Review+Word for large scale production, code on the client side should verify that each word selected by gpt-4o-mini is in the PDT word list. If not, the request to gpt-4o-min (with the same score and word list) should be resent.

\subsubsection{Distribution of Synthetic Sentiment}

\begin{figure}[!htbp]
    \centering
    \includegraphics[width=\linewidth]{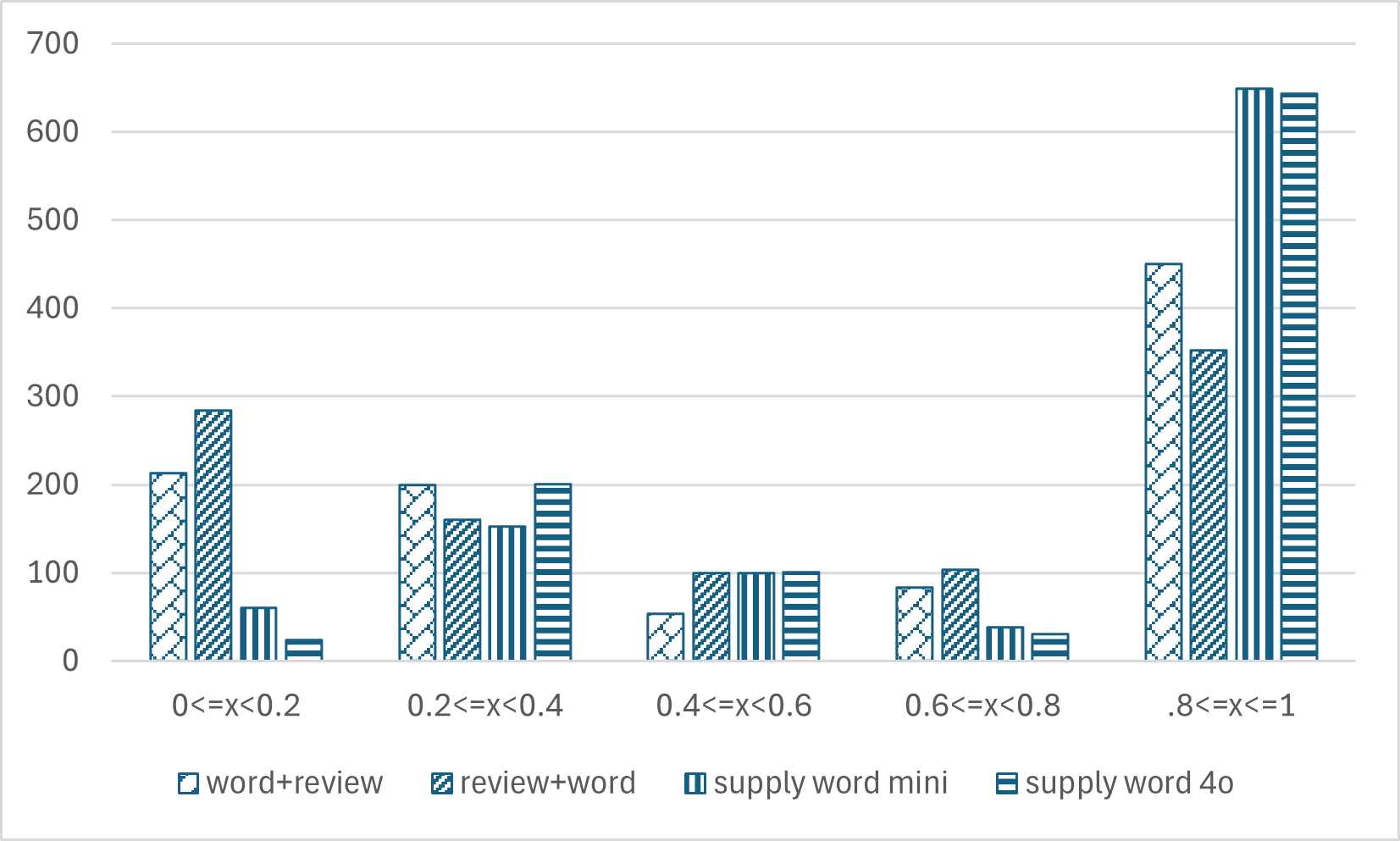}
    \caption{Histogram of Evaluated PDT Scores.}
    \label{fig:histogram}
\end{figure}

Fig. \ref{fig:histogram} shows the distribution of evaluated sentiment scores. Although Word+Review and Review+Word targeted random scores between 0.0 and 1.0, there is clearly bias in the evaluated scores toward the negative and especially positive ends of the range, % (Fig. \ref{fig:histogram}),
with relatively few scores in the mid-range.

The tendency toward the ends of the range is even more pronounced for Supply-Word, whether scored by gpt-4o-mini or gpt-4o. Part of this was caused by the Supply-Word approach, which supplied a word to gpt-4o-mini for scoring. None of these words scored as a 0.0 (unlike Word+Review and Review+Word), resulting in higher target scores and, consequently, higher evaluated scores.

Given the minor differences between Supply-Word-Mini and Supply-Word-4o, one could surmise that the text generation itself is the cause of bias, and not the scoring. Perhaps mixed sentiment is the hardest for gpt-4o-mini to produce, or the model itself is naturally more positive. In addition, the PDT word list itself is inherently skewed with a 60\% positive to 20\% negative and 20\% neutral sentiment word ratio (as acknowledged in the original PDT paper~\cite{Benedek}), so the positive bias in the evaluated scores is not entirely surprising, although the influence of word list bias on evaluated scores is undetermined at this point. Producing a dataset with a more even distribution of evaluated scores (if that is a goal) might simply require weighting the target scores more heavily in certain areas, but that will require further experimentation.

\subsection{Alignment between Target and Evaluated Scores (RQ2)}

\begin{figure*}[!htbp]
    \centering
    \includegraphics[width=\linewidth]{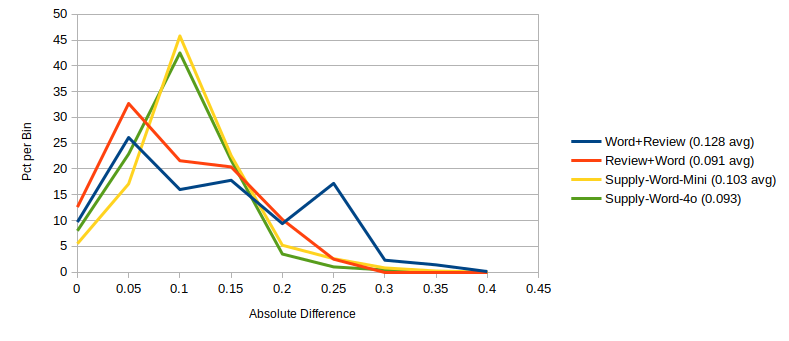}
    \caption{Distribution of Absolute Differences between Target \& Evaluated Scores per Method by Percentage. \textit{Results represent bins 0, 0.01-0.05, 0.06-0.10, ...}}
    \label{fig:diff_per_method}
\end{figure*}

Fig. \ref{fig:diff_per_method} shows the distribution of absolute differences between the target and evaluated scores per method, by percentage of the generated data with an absolute difference value in each bin. The differences were generally low, meaning good alignment and suggesting that gpt-4o-mini can do a reasonably good job of producing synthetic data via any of the described methods. The mean absolute differences for each method were Word+Review (0.128), Review+Word (0.091), and Supply-Word (0.103). 

As shown (in Fig. \ref{fig:diff_per_method}), all methods had low absolute differences, with the supply-word-mini and supply-word-40 having nearly identical distributions. The spike seen around 0.25 for Word+Review  is odd. For the instances in this range, it seemed that gpt-4o-mini was often scoring mixed reviews higher than they otherwise should have been according to the text of the review. An example is Word: `Flexible' and Review: ``I found the software to be quite flexible in adapting to my needs. It allowed me to adjust settings and features to suit my workflow, which made using it much easier. However, sometimes the various options can feel overwhelming until you become accustomed to them. Overall, I'm pleased with how it fits into my routines.'' This hypothetical review was produced in response to a target score of 0.60. The gpt-4o-mini LLM scored this quite high at 0.85, which is arguably more positive than it should have been.

While there were some larger mismatches between target and evaluated scores using gpt-4o-mini (e.g., there were two instances with absolute differences of 0.45 for Supply-Word-Mini), in general the differences were low. In rare cases where the differences were greater, it appeared that gpt-4o-mini got confused in sentiment scoring of long text reviews. Supply-Word-4o did not have this issue as gpt-4o is naturally better at understanding sentiment. However, scoring PDT data at scale would not be cost effective with gpt-4o. Further, Supply-Word-4o is somewhat of an apples to oranges comparison because it compares the target scores of synthetic text generated by gpt-4o-mini (with its own understanding of sentiment) with evaluated scores produced by gpt-4o (with a slightly different understanding of sentiment). While gpt-4o handled the cases where gpt-4o-mini got confused, there were other medium differences across the dataset, presumably because the two models have a slightly different understanding of sentiment. 

Table \ref{tab:alighmentTable} shows the statistical data for the three approaches. Measures include mean, variance, mean absolute difference (MAD), mean squared difference (MSD), Pearson and tStat. As shown, the Pearson correlation results between the target and evaluated sentiment scores highlight the effectiveness of the synthetic data generation methods used in this study. Word+Review, Review+Word, and Supply-Word demonstrated strong correlations, with values of 0.93, 0.96, and 0.97, respectively. These high correlations suggest that gpt-4o-mini can accurately produce product reviews that align closely with the intended sentiment scores. Review+Word and Supply-Word, in particular, exhibited the best alignment, making them more reliable for applications requiring precise sentiment reflection. However, while the high Pearson scores indicate overall success, the slight variations in mean differences, the out of range tStat values, and occasional mismatches in sentiment scoring warrant further exploration to reduce these discrepancies in future work.

\definecolor{lgray}{gray}{0.9}
\definecolor{lgreen}{rgb}{0.78, 0.99, 0.78}

\begin{table*}[ht]
\centering
\caption{Comparison of Target and Evaluated Scores}
\label{tab:alighmentTable}
\begin{tabular}{|l||c|c||c|c||c|c||c|c|}
\hline
 &\multicolumn{2}{c||}{\textbf{Word+Review}} & \multicolumn{2}{c||}{\textbf{Review+Word}} & \multicolumn{2}{c||}{\textbf{Supply-Word-Mini}} & \multicolumn{2}{c|}{\textbf{Supply-Word-4o}} \\ \hline
& \textbf{Expected} &\textbf{Generated}  & \textbf{Expected} &\textbf{Generated}  & \textbf{Expected} & \textbf{Generated} &\textbf{Expected} &\textbf{Generated} \\ \hline\hline
 \textbf{Mean} &0.51  & 0.58 &0.50 & 0.53 & 0.65 &0.74 &  0.65 &0.74 \\ \hline
 \textbf{Variance} &0.09  & 0.13 &0.09 & 0.13 & 0.08 &0.09 &   0.08 &0.09  \\ \hline
\textbf{MAD} &0.13  &  &0.09 &  & 0.10 & &  0.09 &  \\ \hline
\textbf{MSD} &\cellcolor{lgreen}0.02  &  &\cellcolor{lgreen}0.01  &  &\cellcolor{lgreen}0.01 &  & \cellcolor{lgreen}0.01 & \\ \hline
\textbf{Pearson} &\cellcolor{lgreen}0.93  &  &\cellcolor{lgreen}0.96 &  & \cellcolor{lgreen}0.97 & & \cellcolor{lgreen}0.98 &  \\ \hline
\textbf{tStat} &-16.43  &  &-8.14 &  &-43.00 & & -45.90 &  \\ \hline
\end{tabular}
\end{table*}

\subsection{Text diversity scoring (RQ3)}
Table \ref{tab:diversity_scoring} shows the results of text diversity scoring of the three methods along with the number of words in each dataset. Within the Supply-Word dataset, 79.3\% of the reviews started with ``I recently.'' Despite that redundancy, the low CR, CR-POS and HS scores, and high NDS score suggest greater text diversity for that dataset than the others. Likewise, the scores for the Word+Review dataset suggest greater text diversity than the Review+Word dataset. The caveat to these findings is the difference in dataset sizes, which may limit confidence in the meaning found in direct comparisons. More comparable results might be obtained if the datasets were of similar size (according to \cite{shaib2024standardizing}).

\begin{table}[htbp]
\centering
\caption{Text Diversity Scoring for the Three PDT Datasets}
\label{tab:diversity_scoring}
\begin{tabular}{|c|c|c|c|c|c|}
\hline
\textbf{Method} & \textbf{CR} & \textbf{CR-POS} & \textbf{HS} & \textbf{NDS} & \textbf{Words}\\ \hline
Word+Review    & 4.397  & 6.715  & 0.208  & 1.559 & 49,131\\ \hline
Review+Word    & 4.727  & 6.871  & 0.234  & 1.323 & 76,107\\ \hline
Supply-Word         & 3.729  & 6.179  & 0.200  & 1.650 & 128,115\\ \hline
\end{tabular}
\end{table}

\comments{
1. word then review cr= 4.397  cr-pos=6.715  hs= 0.208  nds= 1.559
2. review then word cr= 4.727  cr-pos=6.871  hs= 0.234  nds= 1.323
3. supply word cr= 3.729  cr-pos=6.179  hs= 0.2  nds= 1.65
} 

In terms of practical use, CR, CR-POS and NDS ran quickly. However, HS is quite slow and not recommended for larger datasets. Even for the smaller datasets in this study, HS was quite slow on a basic desktop, taking 41 minutes for the Word+Review dataset, 1 hour 40 minutes for the Review+Word dataset, and 4 hours 43 minutes for the Supply-Word dataset.

\subsection{PDT Dataset Costs (RQ4)}

Table \ref{tab:gen_costs} shows the costs of data generation in terms of time, tokens and gpt-4o-mini price per method. At the time of running the experiments in Oct 2024, gpt-4o-mini dollar costs were relatively low at \$0.15/1M input tokens and \$0.60/1M output tokens~\cite{openAIpricing}, making it more ideal for big data synthesis tasks compared to some of the other massive commercial LLMs. For example, gpt-4o-mini represents a 94\% savings over gpt-4o which costs \$2.50/1M input tokens and \$10.00/1M output tokens~\cite{openAIpricing}. Scaling up generation, the estimated costs to produce a PDT dataset with 1M responses using Word+Review (the cheapest approach) would be $\sim$\$60 with gpt-4o-mini, and $\sim$\$1000 with gpt-4o.

The input sizes (i.e., the prompts sent to gpt-4o-mini) were relatively similar in size. The output from Supply-Word was much more verbose, thus the overall cost for that method was higher. The prompt for Supply-Word (and the others) could likely be adjusted to make the output results less verbose, although longer text might be desired in some situations.

\begin{table}[htbp]
\centering
\caption{Costs for the 1000 Row Synthetic PDT Datasets}
\label{tab:gen_costs}
\begin{tabular}{|c|c|c|c|c|c|}
  \hline
&  & \multicolumn{3}{c|}{\textbf{Tokens}} & \\ \cline{3-5}
\textbf{Method} & \textbf{Time (s)} & \textbf{Input} & \textbf{Output} & \textbf{Total} & \textbf{Price} \\ \hline
Word+Review & 1531 & 126814 & 62462 & 189276 & \$0.06 \\ \hline
Review+Word & 1792 & 105845 & 96189 & 202034 & \$0.07 \\ \hline
Supply-Word & 3084 & 96824  & 161074 & 257898 & \$0.11 \\ \hline
\end{tabular}
\end{table}

The times required to produce the data were relatively high, averaging 1.5-3.1 seconds per review. Although the cost to produce 1M items using Word+Review might be acceptably low, the estimated time to produce such a set for this fastest approach is $\sim$17.72 days which may or may not be acceptable depending on the criticality of such an effort. 

The environmental impacts of training and running AI models are an additional important cost consideration~\cite{strubell2019energy,schwartz2020green,kaack2022aligning}. \citeauthor*{crawford2024generative}~\cite{crawford2024generative} argues that immediate action is needed to limit the environmental impacts of AI. The gpt-4o-mini LLM is designed to be a smaller, more cost efficient model in comparison to larger models~\cite{gpt4omini}, which should normally translate to a reduced carbon footprint. However, OpenAI has not commented specifically on this aspect or the nature of the resources (and thus the associated environmental impacts) consumed in producing gpt-4o-mini relative to other models.

\subsection{Ethical considerations}
The production of synthetic data raises potential ethical considerations. First, because LLMs such as gpt-4o-mini are trained on vast amounts of human text, which inherently contains biases, there will naturally be bias in the generated PDT data. This is easily seen in the words selected in methods 1 and 2, but will be present in the synthetic text as well. This bias might be present in the model's understanding of sentiment. Future efforts could look at options for reducing unintended biases.

Although this is synthetic data, care must be taken to ensure the exclusion of personal identifiable or confidential information, particularly if the data will be produced at scaled and used publicly. The datasets produced in this work were small enough that simple inspection revealed no issues, but an automated approach would be needed at scale.

These synthetic datasets are intended solely for internal sentiment analysis research. If distributing such datasets, they should be clearly marked as synthetic data, and that no real-world decisions (e.g., related to marketing, product development, or public relations) should be made based on the content.

\section{Future work}

This study provided an initial examination of the production of synthetic PDT datasets. 
Future work will investigate potential improvements in a variety of areas including text diversity as well as methods for improving alignment between target and evaluated scores. In addition, we will explore the generation of elements that may appear in actual human datasets (e.g., satire and emojis), along with appropriate measures to assess the human-like quality of such content. Also, we will generate synthetic data for the two software products described in \cite{Hastings2022} and \cite{hastings2002carma}, each of which has corresponding human PDT datasets~\cite{Hastings,Weitl} for which direct comparisons can be made. We will also use the synthetic data to evaluate the method from \cite{weitl2024analyzing}. We also plan to complete a detailed comparison with other existing synthetic data generation tools or methods to test the validity of the methods described in this work.

Because OpenAI does not support persistent session connections, communicating with the GPT models through the API requires resending the entire conversation context with each request. For this study, this meant including the lengthy system prompt, which defined the model's role, in every API call. This greatly increased the overall costs of producing the synthetic PDT data. In the future, we will investigate alternative approaches to having OpenAI generate multiple PDT rows per request to reduce costs.

\section {Conclusion}
This research demonstrates the potential of LLMs, specifically gpt-4o-mini, as a powerful tool for generating synthetic datasets tailored for PDT sentiment analysis. By comparing three methods, Word+Review, Review+Word, and Supply-Word, the study reveals that LLMs can reliably produce synthetic product reviews with strong alignment to target sentiment scores and notable text diversity. While some challenges such as bias toward positive sentiments and occasional misalignment were found, the efficiency and scalability of gpt-4o-mini make it a cost-effective option for large-scale synthetic data generation.

In addition, running simply on a basic desktop computer, the findings suggest that LLMs can be effectively employed in low-resource scenarios, providing valuable alternatives to real-world data acquisition, which can be costly, time-consuming, or even impossible. Future research will focus on refining sentiment alignment techniques, improving textual diversity, and addressing biases inherent in LLMs to further enhance the quality and applicability of synthetic PDT datasets. 
The utilization of LLMs in synthetic PDT data generation not only advances the field of sentiment analysis, but also offers significant opportunities for scaling data-driven product development efforts.

\balance
\printbibliography
\end{document}